%% file: main.tex
\documentclass[runningheads]{llncs}

\usepackage{eccvabbrv}

\usepackage[T1]{fontenc}
\usepackage{graphicx}

\usepackage{color}
\definecolor{myred}{rgb}{.8,.0,.0}

\usepackage{url}
\usepackage[pagebackref]{hyperref}
\usepackage{subcaption}
\usepackage{amsmath,amsfonts}

\ifdefined\DOUBLEBLIND
\newcommand{\repourl}{\url{https://anonymous.4open.science/r/dataset_diversity-03F9}}
\else
\newcommand{\repourl}{\url{https://github.com/TheoSourget/dataset_diversity_evaluation}}
\fi

\begin{document}
\title{Dataset Diversity Metrics and Impact on Classification Models}
\titlerunning{Dataset Diversity Metrics and Impact on Classification Models}
\ifdefined\DOUBLEBLIND
    \author{***}
    \authorrunning{***}
    \institute{***}
\else
    \author{Théo Sourget\inst{1} \and
    Niclas Cla{\ss}en\inst{1} \and
    Jack Junchi Xu\inst{2,3} \and
    Rob van der Goot \inst{1} \and
    Veronika Cheplygina\inst{1} \\
    }
    \authorrunning{T. Sourget et al.}
    \institute{
    IT University of Copenhagen, Denmark\and
    Copenhagen University Hospital, Herlev and Gentofte, Denmark \and
    Radiological AI Testcenter, Denmark \\
    \email{\{tsou,vech\}@itu.dk}\\
    }
\fi
\maketitle              
\begin{abstract} 
The diversity of training datasets is usually perceived as an important aspect to obtain a robust model. However, the definition of diversity is often not defined or differs across papers, and while some metrics exist, the quantification of this diversity is often overlooked when developing new algorithms. In this work, we study the behaviour of multiple dataset diversity metrics for image, text and metadata using MorphoMNIST, a toy dataset with controlled perturbations, and PadChest, a publicly available chest X-ray dataset.
We evaluate whether these metrics correlate with each other but also with the intuition of a clinical expert. We also assess whether they correlate with downstream-task performance and how they impact the training dynamic of the models.
We find limited correlations between the AUC and image or metadata reference-free diversity metrics, but higher correlations with the FID and the semantic diversity metrics. 
Finally, the clinical expert indicates that scanners are the main source of diversity in practice. However, we find that the addition of another scanner to the training set leads to shortcut learning.
The code used in this study is available at \repourl

\end{abstract}

\section{Introduction}
\label{sec:intro}
\input{sec01_intro}

\section{Related Work}
\input{sec02_related}

\section{Methods}
\input{sec03_methods}

\section{Results}
\input{sec04_results}

\section{Conclusive remarks}

\input{sec05_discussion}

\section*{Acknowledgements} 
\ifdefined\DOUBLEBLIND
***
\else
This study was funded by the Novo Nordisk Foundation (grant number NNF24OC00926). 
We want to thank the providers of the MorphoMNIST and PadChest datasets used in this study.
We also thank Inna Ermilova and Ties Robroek for their help with the computational resources and training monitoring.
\fi
%
%
%
\bibliographystyle{splncs04}
\bibliography{refs}

\end{document}

%% file: sec01_intro.tex
\begin{figure}[t]
    \centering
    \includegraphics[width=\linewidth]{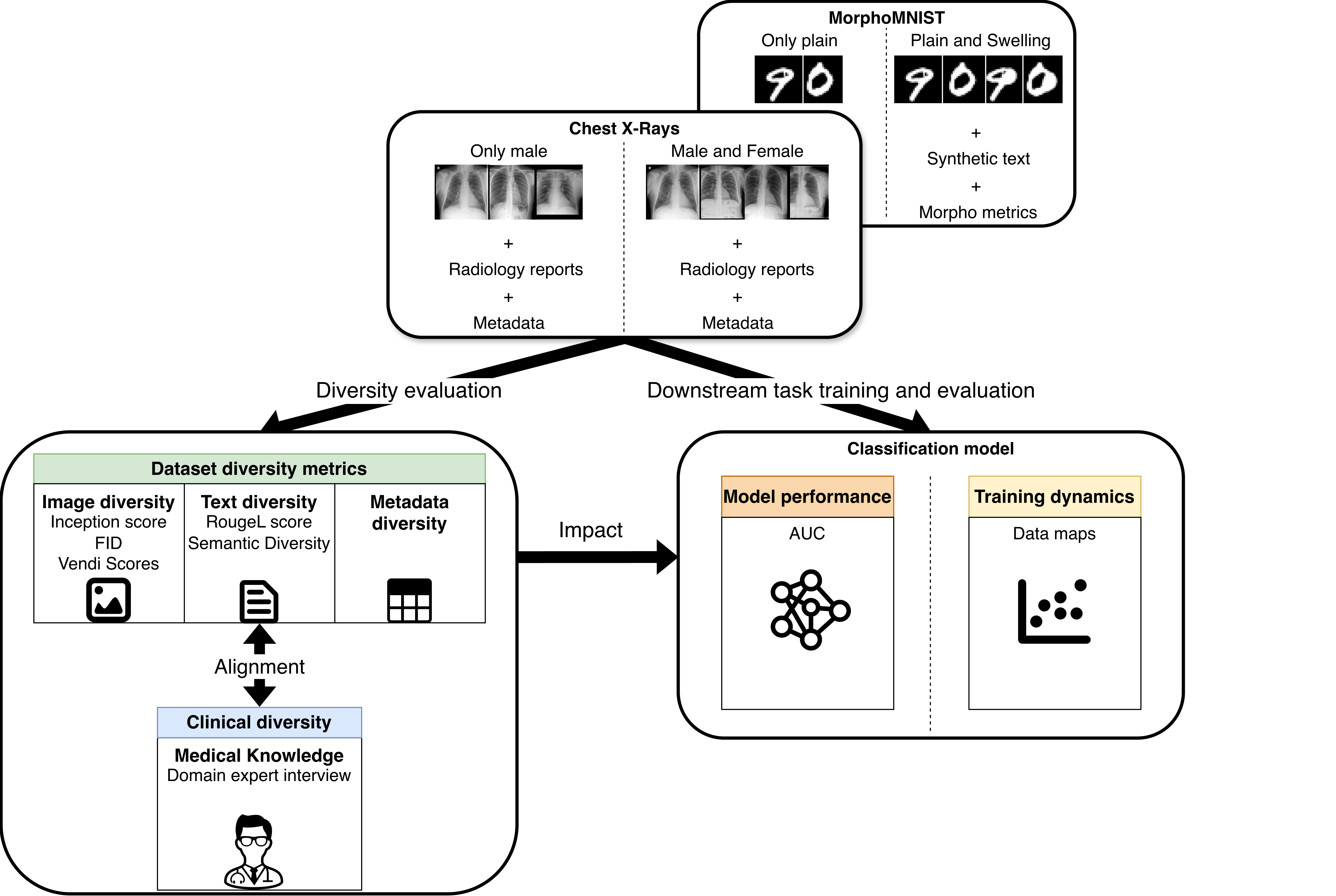}
    \caption{Overview of our study. We assessed dataset diversity measures across multiple modalities and compare their correlations as well as their alignment with domain experts knowledge through interviews. We evaluate their impact on downstream task performances and the effect on subgroups training dynamics.}
    \label{fig:graphical_abstract}
\end{figure}

While diversity is often said to be an important dataset property to train accurate and robust models~\cite{ganapathi2022tackling,jimenezsanchez2025picture}, medical image analysis papers often focus on methodology to improve the downstream performances, overlooking the evaluation of the data used in the studies \cite{varoquaux2022machine}. 
Moreover, the definition of diversity itself is not always clear and may vary across papers~\cite{zhao2024measure}. It can for example relate to the actual content of the data (e.g. the variety of words in text datasets or the visual aspects in image datasets), to the label distribution, or to meta-information such as the demographics of patients.
While medical imaging datasets are often of limited size compared to natural imaging datasets, larger datasets are becoming available~\cite{bustos2020padchest,johnson2019mimic,mei2022radimagenet}. However, the assumption that larger datasets are necessarily more diverse and representative is not always true~\cite{varoquaux2025hype}. In the medical context, the addition of new data may even damage the performance of a classifier, especially in multi-center datasets where inclusion of data from other hospitals may introduce a distribution shift or a potential shortcut~\cite{compton2023when,shen2024data}. The increasing size of datasets and sometimes their curation from online sources~\cite{ikezogwo2023quilt} have also been shown to introduce quality problems~\cite{aubreville2024model}.  

Several metrics exist to measure the diversity of data, often used in the context of generative models to ensure that they produce diverse set of samples~\cite{heusel2017gans,kynkaeaenniemi2019improved,salimans2016improved}. These metrics are yet to be evaluated on medical image classification datasets. Moreover, it is still unclear whether an improvement on these diversity metrics actually translates to better downstream performances or if they align with clinicians' interpretation of diversity. Finally, most studies focus on evaluating diversity of a single modality, which is not sufficient for increasingly popular multimodal datasets~\cite{bustos2020padchest,johnson2019mimic}.
Our contributions, aiming for a better understanding of diversity metrics and their impact on model trainings in various applied scenarios, are as follows:
\begin{enumerate}
    \item We assess image, text, and metadata diversity metrics on a toy synthetic dataset and on chest X-ray data, providing insights into their strengths and weaknesses in different scenarios. 
    \item We evaluate their impact on models' performance, fairness, robustness, and training dynamics, finding which metrics are suitable for the selection of a training set.
    \item We conduct a semi-structured interview with a radiology resident to gather the clinical view of diversity and discuss its alignment with diversity metrics and research practices.
\end{enumerate}

%% file: sec02_related.tex
There are multiple definitions of dataset diversity, yet it is not always clear which definition is used when a paper addresses diversity. In their taxonomy, Zhao \etal~\cite{zhao2024measure} identify five definitions said to increase diversity: (1) the content of the data (e.g. variety of objects in images or words in texts), (2) the location of the data collection, (3) the domain of the data (e.g. type of images or texts), 4) the demographics within the dataset, and 5) the background of the annotators. Closely related to diversity, Clemmensen and Kjærsgaard~\cite{clemmensen2022data} discuss how representativity of datasets is described in research papers. They similarly find that representativity is not always clearly defined, and sometimes simply claimed without supporting arguments. In our work, we choose to explicitly focus on the diversity in the content of the data and the demographics within the dataset.

Other studies have focused on the measures for dataset quality. Mitchell \etal~\cite{mitchell2022measuring} detail metrics to quantify five different aspects of machine learning datasets: (1) distance, (2) density, (3) diversity, (4) tendency, and (5) association. They categorize the metrics as ``General'' (e.g. Gini diversity and Vendi Score) or ``Modality-specific'' (e.g. lexical diversity, Inception Score, and subset diversity).
Dang \etal~\cite{dang2024data} also present a taxonomy for data quality and metrics for natural language processing (NLP) datasets, including diversity, defined as the ``variety and range covered by the dataset''. They compare five metrics: the Shannon entropy, the Renyi entropy, the Simpsons index, the Vendi Score, and the inter-class distance introduced in their study.

Diversity is considered an important characteristic of generative models outputs to match the entire training distribution and avoid the generation of identical samples (also referred to as mode collapse). Multiple works address its quantification ~\cite{albuquerque2025benchmarking,dombrowski2025image,ibarrola2024measuring,raisa2025position}, Dombrowski \etal~\cite{dombrowski2025image} also introduce a new metric, the Image Retrieval Score as the number of real images being the closest to at least one of the synthetic image. 

Closely related to our study, Mironov \etal~\cite{mironov2025measuring} highlights the limitations of existing diversity measures using simple failures cases. They define three desirable properties for a diversity metric: monotonicity, uniqueness, and continuity. They categorize existing metrics according to whether they satisfy these properties and propose two metrics that satisfy all criteria, but which are not usable in practice due to their computational cost. 

Finally, Konz \etal~\cite{konz2026frechet} show the limitations of existing measures to compare two sets of medical images and present the Fréchet Radiomic Distance (FRD) using clinically meaningful features within images. They use the FRD to evaluate image-to-image translation models on a large range of medical downstream tasks, and show that the FRD is better correlated with the downstream task metric like the AUC or the Dice score.

We extend these works by applying diversity metrics across different modalities (image, text, and tabular) in the context of medical classification datasets, providing insights into their behaviour, advantages and failure cases when applied to real images. 
Moreover, beyond studying the behaviour of the diversity metrics, we measure if they correlate with a change in downstream-task performance and in training dynamics.
Finally, we leverage the application to a real-world scenario to assess whether the metrics relate to domain expert definition of diversity.

%% file: sec03_methods.tex
\subsection{Problem Setting}

In our supervised classification setting, given a multimodal dataset \(D_{eval} =\{x,y\}^N_{n=1}\) with \(x_n = (img_n,txt_n,metadata_n)\), we quantify the diversity \(Div_{D_{eval}}\) of this dataset using a diversity metric function \(f\). We distinguish two main types of metric \(f\) based on the need for a reference dataset \(D_{ref}\):
\begin{enumerate}
    \item Reference-free metrics where \(Div_{D_{eval}} = f(D_{eval})\)
    \item Reference-based metrics where \(Div_{D_{eval}} = f(D_{eval},D_{ref})\)
\end{enumerate}

Given a set of scenarios (options to select a subset of the dataset), our goal is to evaluate how each diversity measure ranks the scenarios such that the highest ranked is correlated with better performance on a downstream classification task when used to train a classifier.

\subsection{Data}
\subsubsection{Toy dataset:}
\begin{figure}[t]
    \centering
    \includegraphics[width=0.80\linewidth]{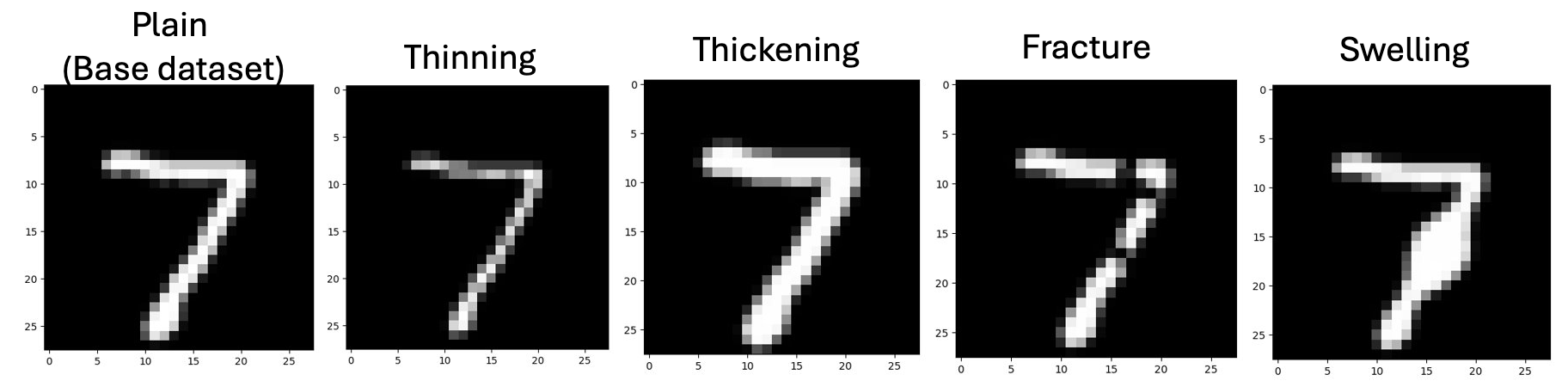}
    \caption{Example of data from MorphoMNIST}
    \label{fig:MorphoMNIST}
\end{figure}

We use a toy dataset to have a better control on the data and on the diversity. MorphoMNIST\footnote{Data downloaded from \url{https://drive.google.com/uc?export=download&id=1-E3sbKtzN8NGNefUdky2NVniW1fAa5ZG}}~\cite{castro2019morpho} is a framework extending the MNIST dataset~\cite{deng2012mnist} with controlled shape perturbations: thinning, thickening, fractures, and swelling (see Fig.~\ref{fig:MorphoMNIST} for examples). We use the train and test splits from the original MNIST dataset. The train split is used as the dataset to be evaluated by the metrics as well as the data to train classifiers for digit classification. The test split is used as the reference set for the FID and to compute the performance metric on the digit classification task.
The testing set is fixed and contains an equal amount of images for all perturbations (50,000 images in total, 10,000 images per perturbations). We generate multiple training sets with varying perturbations creating scenarios either containing a single type of image or containing both plain digits and digits with a perturbation: (1) Plain, (2) Thin, (3) Thick, (4) Fracture, (5) Swelling, (6) Plain $\cup$ Thin, (7) Plain $\cup$ Thick, (8) Plain $\cup$ Fracture, and (9) Plain $\cup$ Swelling.

Inspired by the setup of~\cite{schrodi2025two}, we generate synthetic text for each sample using the label and the applied transformation (e.g. ``Image of a handwritten plain seven'' or ``Image of a handwritten thin nine''). 

\subsubsection{Chest X-ray dataset:}
We use the publicly available dataset PadChest\footnote{Data accessed from \url{https://bimcv.cipf.es/bimcv-projects/padchest/.}}~\cite{bustos2020padchest} to conduct our experiments on chest X-rays.

\begin{figure}[t]
    \centering
    \includegraphics[width=0.5\linewidth]{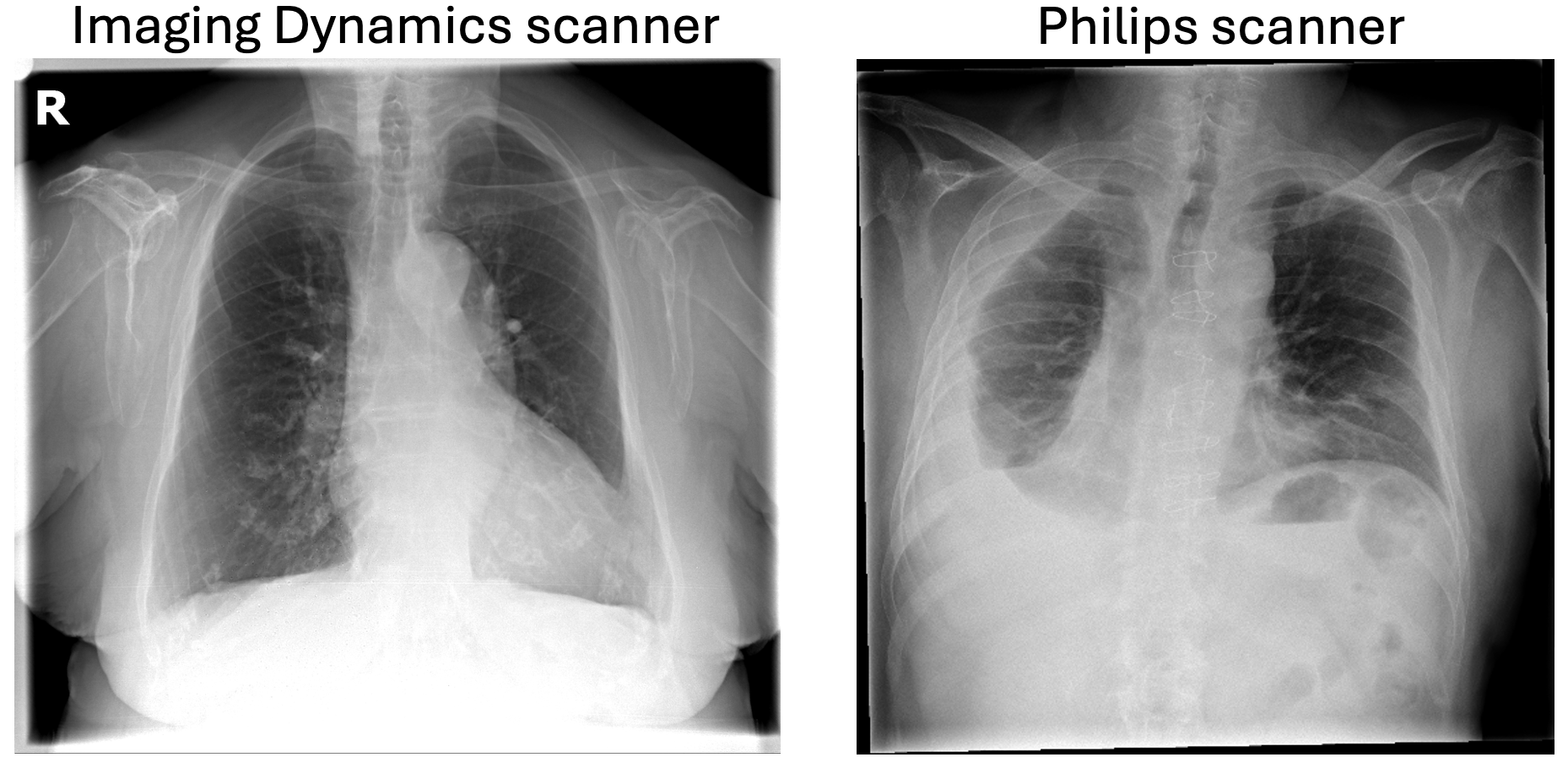}
    \caption{Example of data from both scanners in the PadChest dataset.}
    \label{fig:PadChest}
\end{figure}

PadChest contains more than 160,000 chest X-rays acquired using two scanners (see Fig.~\ref{fig:PadChest} for examples of images from both scanners), Spanish radiology reports, and metadata (e.g. patient sex, patient age, and scanner used for the acquisition) from the San Juan Hospital. We removed images of lateral views, images for which the label was null or included ``suboptimal study'', ``exclude'' or ``unchanged'', and entirely black images, resulting in a final set of 108,076 images. The dataset provides annotations for 174 labels but in this study we only focus on the pneumothorax label. No training and testing sets are provided by the dataset, we therefore create a testing split with 20\% of the full dataset, stratified on the class label as commonly done in training pipelines. We ensure that all images for a patient are in the same split as it is important to avoid data leakage in medical image analysis tasks~\cite{rumala2023how}.

We create five scenarios to evaluate diversity: (1) All using the entire dataset, (2+3) Female and Male using only patients of that sex, and (4+5), Philips and ImagingDynamics using only images acquired with that scanner.

\subsection{Diversity Metrics}
We compare the correlation between diversity metrics, evaluating whether they behave similarly across scenarios. For each metric, we rank the different scenarios of a dataset and compute the correlation between the rankings of two metrics using the Spearman's rank correlation~\cite{zar2005spearman}. We use the rankings instead of the actual values as we believe diversity metrics' objective is to determine the best set and therefore the ranking is the most valuable information.

For PadChest, we also use the bootstrap method~\cite{efron1992bootstrap} with 10 resamplings to estimate the 95\%-confidence intervals. We perform the bootstrap with few resamplings due to the computational expensiveness of some metrics, reducing the robustness of the confidence interval.

We assess the following metrics grouped by the modality they are evaluating:

\subsubsection{For images:}
1) The Inception Score (IS)~\cite{salimans2016improved}: originally defined to evaluate generative models, the IS evaluates both the quality and diversity of images using an InceptionV3 model. The metric is maximized when the entropy of the conditional probability distribution is low (probabilities are close to a one-hot vector, meaning the classes are distinct), and when the entropy of the marginal probability distribution is high (the prediction of the model is similar across classes). This is calculated using the Kullback-Leibler divergence between the conditional probability distribution and the marginal probability distribution: 
\begin{align}
\text{IS} = \exp(\mathbb{E}_x\text{KL}(p(y|x)||p(y)))
\end{align} 
2) The Fréchet Inception Distance (FID)~\cite{heusel2017gans}: also used in the evaluation of generative models, the FID requires another dataset used as a reference. The FID compares the mean and standard deviation of the InceptionV3 features generated with the two datasets:
\begin{align}
\text{FID} = ||\mu_{ref}-\mu_{eval}||^2+T_r(\Sigma_{ref}+\Sigma_{eval}+2(\Sigma_{ref}\Sigma_{eval})^{1/2})
\end{align} 
3) The Vendi Score (VS)~\cite{friedman2023vendi}: derived from an ecological definition of diversity as the exponential of the
entropy of the distribution of species, the Vendi Score extend this definition to ML settings. Relying on a positive semidefinite similarity function between samples of a dataset, this metric can therefore be applied to any modality, and doesn't need a reference dataset or a model to be computed. The Vendi Score is define as the exponential of the Shannon entropy of the eigenvalues \(\lambda\) of the similarity matrix:
\begin{align} 
\text{VS}=\exp(-\sum_{i=0}^{n}\lambda_i\log\lambda_i)
\end{align} 
We use the cosine similarity function with three types of feature vectors. To avoid the reliance on a model in the Inception Score and the Fréchet Inception distance we first use the pixel values as feature vector (referred as VS\_pixel). Still avoiding the usage of a pre-trained model, we also use the histogram of oriented gradients~\cite{dalal2005histograms} (VS\_hog). Finally, to have a closer comparison with the FID and IS, we use the features from the InceptionV3 model (VS\_inception). 

As the Vendi Score is computationally expensive, we need to compute it on a downsampled dataset. We therefore select 10\% of the data in a stratified way to preserve the class distribution and compute the Vendi Score. We repeat this process five times and keep the mean value across the five Vendi Scores obtained.

\subsubsection{For text:}
For text-diversity, inspired by~\cite{zhang2025verbalized}, we evaluate both the lexical and semantic diversity. 

(1) The lexical diversity is computed using the mean RougeL F1-score~\cite{lin2004rouge} between all pairs of text. The RougeL F1-score computes the longest common subsequence between two sentences, a low RougeL score therefore indicates a low overlapping between the two sentences and a higher diversity. Similarly to the Vendi Score with a distance between all pairs of sample, we downsample 10\% of the data in a stratified way to handle the computational complexity of the metric and report the mean of five selections.

(2) The semantic diversity is the mean cosine similarity between all pairs of embeddings generated by a BGE-M3 model\footnote{Accessed via \url{https://huggingface.co/BAAI/bge-m3}}~\cite{chen2024m3}. We chose the BGE-M3 model for its capacity to work with short and long texts and with texts from multiple languages.

\subsubsection{For metadata:}
We compute the metadata diversity using the mean cosine similarity between samples. For MorphoMNIST we use the morphometrics provided by the MorphoMNIST framework (area, length, thickness, slant, width, and height of the digit).  
For PadChest, we use the age, sex, projection, and scanner. Missing information is embedded as -1 in the metadata vector. 

\subsection{Training and Evaluation on Downstream Task Performances}
We compare the performances of models trained with the different scenarios of the datasets. 
We train ResNet-50 models~\cite{he2016deep} with randomly initialized weights, using the same set of hyperparameters across all versions of training datasets and the Adam optimizer~\cite{kingma2015adam} with a learning rate of 1e-4 for a maximum of 100 epochs with early stopping on the validation AUC.  
We use the cross entropy loss for MorphoMNIST and the weighted cross entropy for PadChest to account for the large class imbalance. The weight of the positive samples is therefore defined as the ratio between the number of negative samples (healthy patients) and the ratio of positive samples (patients with pneumothorax).
We log training performances using the Resource Aware Data science Tracker~\cite{robroek2023data}.

We evaluate the models using the AUC as recommended for medical imaging classification~\cite{maierhein2024metrics}. We use a stratified 5-fold splitting to evaluate the variability of the models and report the standard deviation of the five models on the test set. In case of datasets with multiple versions of the same digit (e.g. Plain $\cup$ Thin) or for patients with multiple images, we ensure all images are put in the same split to avoid data leakage.

To complement the performance analysis on fully trained models, we also study the training dynamics across the different scenarios using data maps~\cite{swayamdipta2020dataset}. Data maps track the evolution of the probability in the correct class for each data sample across training epochs. While the original paper uses it on the training samples, we are interested in the evolution on the test samples. We study the evolution of the test samples to compare the dynamics of subgroups when they are present or absent of the training set. Each sample in the map is represented using two statistics: 1) the confidence as the mean probability of the class across epochs, and 2) the variability as the standard deviation of the probabilities across epochs. A sample with a high confidence and low variability (top-left part of the data map) is therefore more easily learnt by the model during training.
A high number of samples may reduce the readability of data maps due to the overlapping of points. In such cases, we report kernel density plots instead of the original maps presented by Swayamdipta \etal~\cite{swayamdipta2020dataset}.

\subsection{Diversity in Clinical Practice}
Finally, we compare whether the behavior of diversity metrics follows the intuition of domain experts. To this end, we conduct a semi-structured interview with a radiology resident asking them to comment on their vision of diverse patients and different scenarios (e.g. are chest X-rays for male and female patients different in the context of pneumothorax). Our goal is to compare if the clinical diversity and the technical diversity researchers may focus on in studies are aligned.

The interview was structured around the following guiding questions: (1) What do you think about when you hear that a collection of chest X-rays is diverse?, (2) How is diversity addressed in your learning as a radiologist?, (3) Would it make sense to ask a radiologist to compare diversity of different subgroups? and (4) From the metadata available in the PadChest dataset, which one would be relevant for diversity and could you think of some missing?

%% file: sec04_results.tex
\subsection{Limited Correlation Between Reference-free Diversity Metrics and Classification Performances}
\begin{figure}[t]
    \centering
    \begin{subfigure}{0.49\columnwidth}
        \centering
         \includegraphics[width=\linewidth]{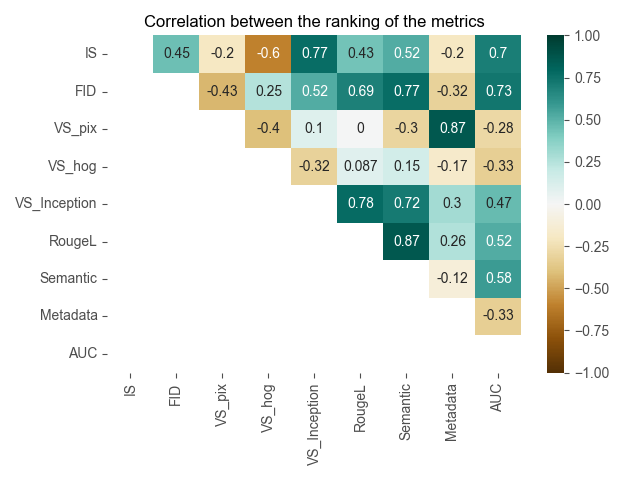}
      \caption{MorphoMNIST}
      \label{subfig:morphomnist_corr}
    \end{subfigure}
    \begin{subfigure}{0.49\columnwidth}
      \centering
      \includegraphics[width=\linewidth]{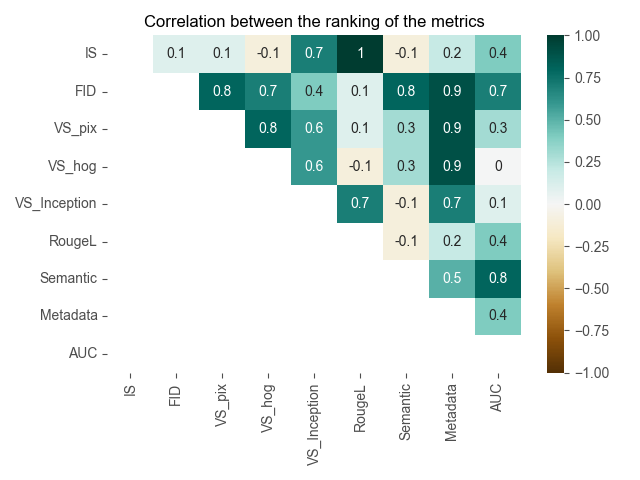}
            \caption{Chest X-rays}
        \label{subfig:cxr_corr}
    \end{subfigure}
    \caption{Correlations between the metrics' ranking of (a) MorphoMNIST and (b) chest X-rays scenarios using the Spearman's rank correlation.}
    \label{fig:metrics_corr}
\end{figure}

Tables~\ref{tab:morphomnist_metrics} and \ref{tab:cxr_metrics} show the metrics for each dataset and scenarios, Fig.~\ref{fig:metrics_corr} shows the correlations of the ranking of the different scenarios by each metric for both datasets.

For MorphoMNIST, while it is a simple toy example we can see that for the metrics based on the Inception models (IS, FID, VS\_inception), the scenarios with plain images and a local perturbation (Fracture or Swelling) leads to the better diversity metrics values. 
It also leads to the best mean AUC over the 5-fold models, however, the digit classification task being simple, all training datasets obtained results close to 0.99 AUC. We can see that the different Vendi Scores have a poor correlation, showing the high impact of the similarity function and how, even for a similar modality, the metric may vary. 
We can also see a correlation between image metrics using the Inception model especially with the Vendi Score and the IS, but poor correlation with the AUC and the Vendi Scores using pixel values and histogram of oriented gradients.
We also notice that the two Vendi Scores relying on non-model based features show a small negative correlation with the AUC.
For the text modality, we can see a strong correlation between the two metrics which was expected due to the simplicity and synthetic aspect of the text data for MorphoMNIST. They also obtain a moderate correlation with the AUC (0.52 and 0.58).

\input{table_metrics_morphomnist}

For PadChest, we observe higher correlations between the different Vendi Scores and the FID. We also observe a strong correlation between these metrics and the metadata diversity ($\geq$0.7). The Inception Score shows low correlation with the other metrics except the Vendi Score using the Inception features and the RougeL. However, aside the FID, the image and metadata diversity metrics do not correlate with the AUC ($\leq$0.4).
Interestingly for the FID, the gap between the entire set diversity and the scanners subgroups diversity is larger than the gap with Male and Female subgroups diversity, which aligns with the radiology resident impressions that scanner is the main source of diversity (see Section~\ref{sec:clinical_diversity} about clinical diversity) and correlates with a large drop of AUC when training on a single scanner.
While the text diversity metrics are very close across configurations and the differences are not visible with two decimal points, the rankings resulting from the slight variations in the semantic diversity is the one with the highest correlation with the AUCs (0.8).

In conclusion, we can see that image and metadata reference-free diversity metrics may not be appropriate to choose the training set leading to the best classification performance. In comparison, the FID showed better correlations but requires the testing set as a reference. Finally, while promising, the narrow range of values for the semantic diversity requires further experiments to verify its relevancy.

\input{table_metrics_cxr}

\input{table_fairness_padchest}

\subsection{Training Dynamics Can Help Detect Shortcuts}
We also study the dynamic of models during training with data maps per type of images in the test set across different training scenarios.

\begin{figure}[t]
    \centering
    \begin{subfigure}{\columnwidth}
        \centering
         \includegraphics[width=\linewidth]{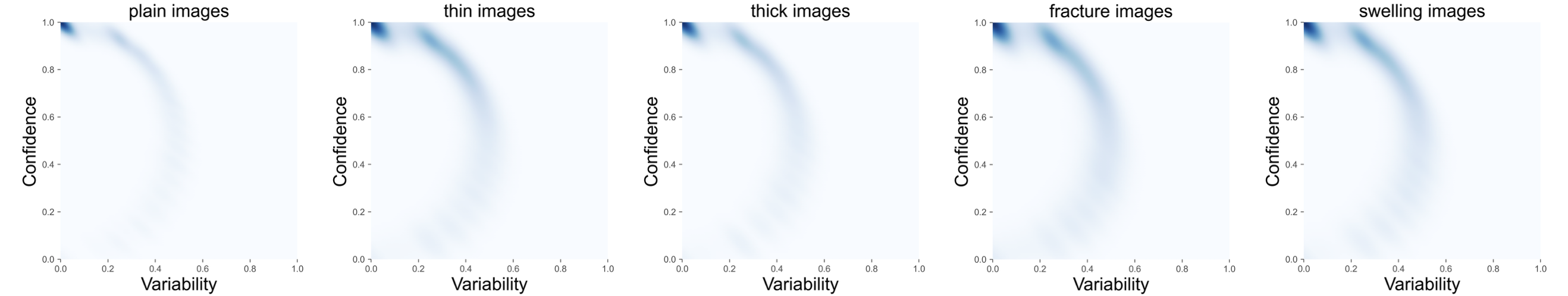}
      \caption{Trained with plain images}
      \label{subfig:datamap_morphomnist_plain}
    \end{subfigure}
    \begin{subfigure}{\columnwidth}
      \centering
      \includegraphics[width=\linewidth]{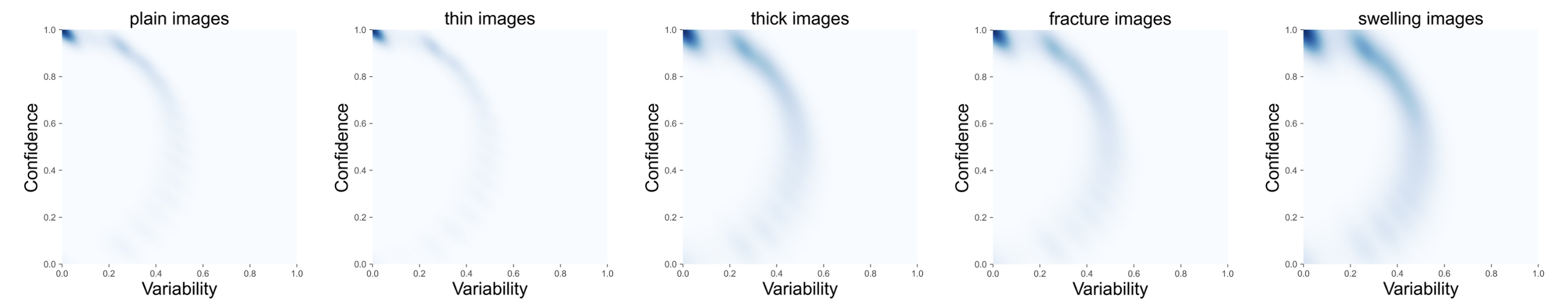}
        \caption{Trained with plain and thin images}
        \label{subfig:datamap_morphomnist_plainthin}
    \end{subfigure}
    \begin{subfigure}{\columnwidth}
      \centering
      \includegraphics[width=\linewidth]{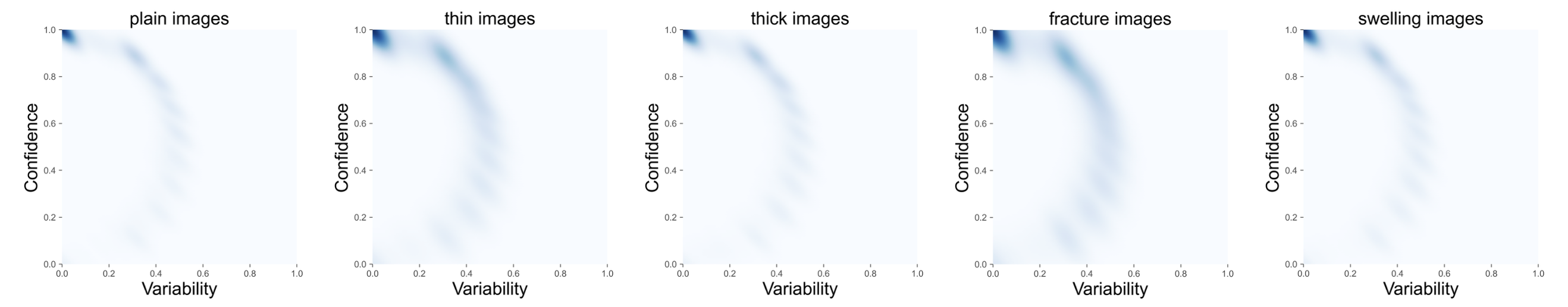}
        \caption{Trained with plain and swelling images}
        \label{subfig:datamap_morphomnist_plainswell}
    \end{subfigure}
    \caption{Density plots of data maps for models trained with (a) plain images only, (b) plain and thin images, and (c) plain and swelling images for the MorphoMNIST dataset. A type of image is learnt faster if the density of the top left part is higher meaning that the probability in the correct class of this type of image is higher and more constant during the training.}
    \label{fig:datamap_morphomnist}
\end{figure}

For MorphoMNIST, despite the simplicity of the task and the small differences between the AUCs of fully trained models, we observe some differences in the training dynamics based on the composition of the training set and the type of perturbation. We observe in Fig.~\ref{fig:datamap_morphomnist} that, as expected, training on a single type of images makes the learning of the images of other perturbations more difficult. However, we can also see differences when trained with only plain images between local perturbation (swelling and fracture) and global ones (thinning and thickening) as the former are more impacted when outside of the training set.
For training sets with multiple types of images we also see a difference in training dynamics between the different combinations. Using Plain $\cup$ Swelling images results in improvements for all types of perturbations, but with less improvement on fractured images. On the other hand, Plain $\cup$ Thin images mostly improved the dynamic of this type of image.
We do not observe clear patterns between the diversity metrics and the training dynamics. The best match is with the FID as the combination Plain $\cup$ Thin images results in a higher value than the other combinations. However the combination Plain $\cup$ Fracture images obtains a better FID than Plain $\cup$ Swelling which does not match the data maps.

\begin{figure}[t]
    \centering
    \begin{subfigure}{\columnwidth}
        \centering
         \includegraphics[width=\linewidth]{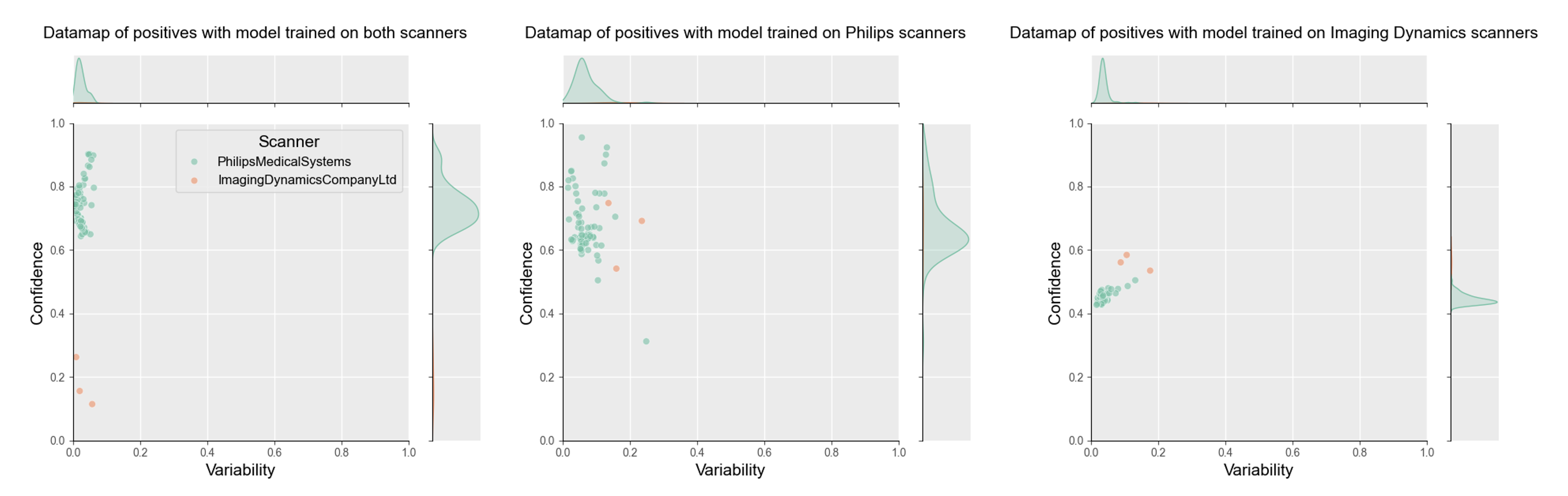}
      \caption{Scanners}
      \label{subfig:datamap_cxr_scanners}
    \end{subfigure}
    \begin{subfigure}{\columnwidth}
      \centering
      \includegraphics[width=\linewidth]{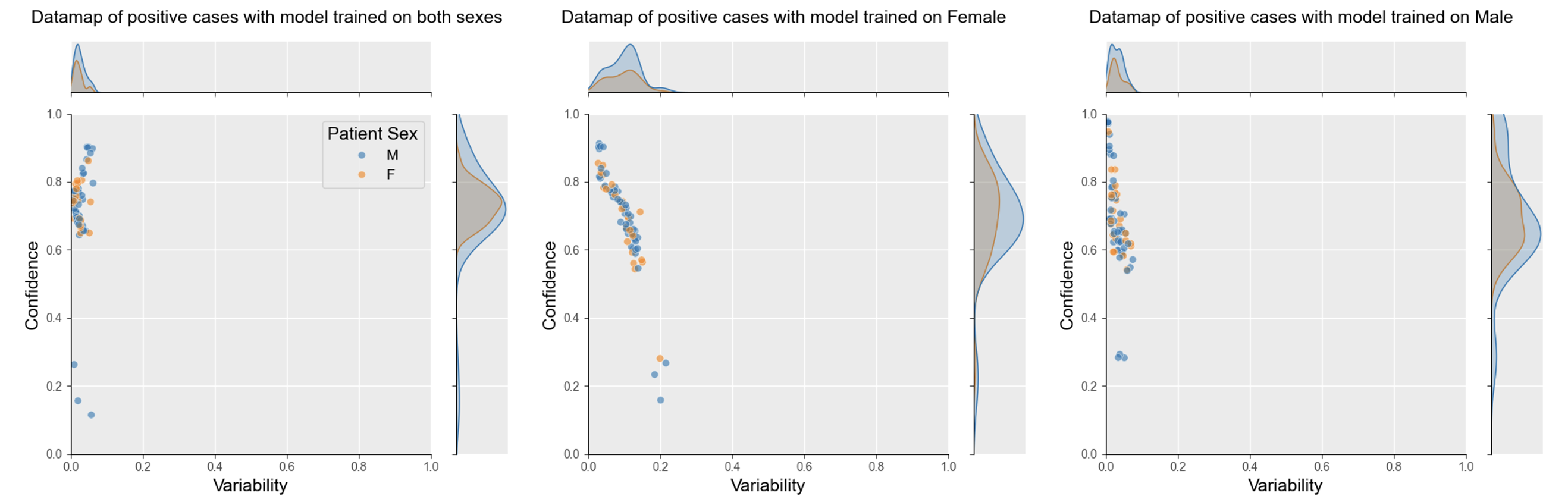}
        \caption{Patient sex}
        \label{subfig:datamap_cxr_sex}
    \end{subfigure}
    \caption{Data maps of positive cases of pneumothorax in the test set of PadChest. Each subplot shows for all samples in the test set the mean and standard deviation of the predicted probability when trained with (a) different scanners, and (b) different patient sex.}
    \label{fig:datamap_cxr_positive}
\end{figure}

For PadChest, we observe in Fig.~\ref{fig:datamap_cxr_positive} with the models trained on the entire dataset, that three samples of pneumothorax cases clearly stand out with a lower mean probability during training. All three samples are the images acquired with the Imaging Dynamics scanner, the scanner with the highest class imbalance. 
Interestingly, when trained with only images from the Phillips scanner, two samples also stand out but they are both acquired with a Phillips scanner as well. Therefore, while completely absent from the training samples, the positive samples from the Imaging Dynamics scanner are better classified. We also observe this clear separation between the two scanners for negative patients with a larger sample size.
This result show the capacity of data maps to guide towards potential shortcuts and that adding data from a new scanner in the training set may not always be beneficial for the robustness of the model. This finding demonstrates further how deep learning models can rely on non-clinically relevant information to classify a disease, as for example shown by Sourget \etal~\cite{sourget2025mask} with models obtaining good performances on images without lungs from the PadChest dataset while it was impossible for a clinical expert to diagnose them. 

With regards to the patient sexes, we can see that the three samples are also only Male patients. However, the same three samples have similar dynamics across the different training sets, even when training only on Male patients. Moreover, we don't observe a gap between the AUCs of Male and Female patients subgroups for the different models in Table ~\ref{tab:cxr_fairness}. We however notice that the models trained only on Male patients obtain better performances on both subgroups compared to the models trained with all patients or only with Female patients.

\subsection{Clinical Diversity}\label{sec:clinical_diversity}
\subsubsection{Scanners as main source of diversity.}
When asked about their view of diversity, the radiology resident pointed the scanner used to acquire the image as the main aspect. Due to the different reconstruction algorithms applied across manufacturers or scanners, the X-ray images texture will appear different to the radiologist. 
Interestingly, information on the reconstruction algorithm is not always retrievable in datasets, for instance in the PadChest dataset only the scanner manufacturer is indicated but the algorithm may vary for different devices of the same manufacturer.
Additionally, it is worth noting that the X-rays quality in the clinical setting may not always be perfect (e.g. part of the lung missing or presence of foreign objects).

\subsubsection{Patient demographics used as statistical guidance.}
While patient demographics was initially not mentioned by the radiology resident in the source of diversity, they indicated that they do learn about differences between male and female patients during their training. However, this difference is not on the disease characteristics in the image but rather on the prevalence of this disease between the two populations. This information may therefore guide the diagnosis, especially in cases where two diseases have similar radiological patterns but one is more prevalent in the patient's population.
Similarly, while ethnicity is generally not a topic widely addressed during training, the geographic location of a patient can also be useful to guide the diagnosis. It is important to note that while previous works have shown the capacity of machine learning models to classify the ethnicity of a patient based on a chest X-ray~\cite{gichoya2022ai} which is not possible for a radiologist.

\subsubsection{Disease severity is not important for chest X-ray data.}
Although our intuition is that having different severities for a disease may be essential for a dataset in order to have a broader view of its visual aspect, it doesn't appear as an important feature to the radiologist when using chest X-ray data. They explain that chest X-rays may not be precise enough to capture small differences or small findings (e.g. tumors with a size inferior to 2cm). Therefore, chest X-rays are generally used to increase or decrease the likelihood of diagnoses rather than assessing the severity. The acquisition or reporting of different severities may therefore not be the most important feature in most cases when curating a chest X-ray dataset compared to CT scans for which the differences are more visible. On the other hand, the radiologist pointed out the importance of measuring the performances when multiple diseases co-occurs, especially for algorithms not focusing on a single disease but on a general diagnosis.

%% file: table_metrics_morphomnist.tex
\begin{table}[htb]
\centering
\caption{Metrics on the MorphoMNIST dataset. ↑ indicates that a higher value is best, ↓ indicates that a lower value is best. +/- for AUC scores indicate the standard deviation of the five models trained using 5-fold cross-validation on the test set. N indicates the number of samples in the set.}
\label{tab:morphomnist_metrics}
\resizebox{\textwidth}{!}{%
\begin{tabular}{l|r|r|r|r|r||r|r|r|r|}
				 & \multicolumn{1}{|c|}{Plain} & \multicolumn{1}{|c|}{Thin} & \multicolumn{1}{|c|}{Thick} & \multicolumn{1}{|c|}{Fracture} & \multicolumn{1}{|c||}{Swelling} & \multicolumn{1}{|c|}{Plain $\cup$ Thin} & \multicolumn{1}{|c|}{Plain $\cup$ Thick} & \multicolumn{1}{|c|}{Plain $\cup$ Fracture} & \multicolumn{1}{|c|}{Plain $\cup$ Swelling} \\
				 & \multicolumn{1}{|c|}{(N=60k)} & \multicolumn{1}{|c|}{(N=60k)} & \multicolumn{1}{|c|}{(N=60k)} & \multicolumn{1}{|c|}{(N=60k)} & \multicolumn{1}{|c||}{(N=60k)} & \multicolumn{1}{|c|}{(N=120k)} & \multicolumn{1}{|c|}{(N=120k)} & \multicolumn{1}{|c|}{(N=120k)} & \multicolumn{1}{|c|}{(N=120k)} \\

\hline
IS ↑	&2.28	&2.24	&2.35	&2.39	&2.36	&2.36	&2.35	&\textbf{2.41}	&2.40	\\
FID ↓	&12.26	&31.70	&13.66	&19.82	&25.22	&14.72	&9.33	&\textbf{5.49}	&6.24	\\
Vendi (pixel values) ↑	&23.64	&\textbf{38.23}	&13.89	&15.89	&30.77	&30.45	&18.45	&20.21	&27.25	\\
Vendi (hog) ↑	&9.70	&9.66	&\textbf{10.20}	&9.10	&8.44	&9.84	&10.09	&9.48	&9.25	\\
Vendi (Inception) ↑	&5.19	&5.45	&5.55	&5.52	&5.55	&5.62	&5.55	&5.81	&\textbf{5.94}	\\
\hline
RougeL (lexical) ↓	&0.85	&0.85	&0.85	&0.85	&0.85	&\textbf{0.76}	&\textbf{0.76}	&\textbf{0.76}	&\textbf{0.76}	\\
Semantic diversity ↓	&0.88	&0.88	&0.87	&0.87	&0.9	&0.82	&0.82	&0.81	&\textbf{0.80}	\\
\hline
Metadata diversity ↓	&0.98	&0.98	&0.98	&0.99	&0.98	&\textbf{0.97}	&0.98	&0.98	&0.98	\\
\hline
AUC ↑	&0.992	&0.989	&0.989	&0.993	&0.990	&0.991	&0.991	&0.993	&\textbf{0.994}	\\
&$\pm$0.002	&$\pm$0.003	&0.002	&$\pm$0.001	&$\pm$0.003	&$\pm$0.003	&$\pm$0.001	&$\pm$0.000	&$\pm$0.000	\\
\hline
\end{tabular}
    }
\end{table}

%% file: table_metrics_cxr.tex
\begin{table}[htb]
\centering
\caption{Metrics on the chest X-ray dataset. ↑ indicates that a higher value is best, ↓ indicates that a lower value is best. Values in [] indicate the 95\% confidence interval computed with the bootstrap method. $\pm$ for AUC scores indicate the standard deviation of the five models trained using 5-fold cross-validation on the test set. We show two decimal points for each confidence interval which is why some confidence intervals appear to have an equal lower and upper bound. N indicates the number of samples in the set and P the number of positive samples.}
\label{tab:cxr_metrics}
\resizebox{\textwidth}{!}{%
\begin{tabular}{l|c||c|c||c|c|}
				 & \multicolumn{1}{|c||}{All} & \multicolumn{1}{|c|}{Female} & \multicolumn{1}{|c||}{Male} & \multicolumn{1}{|c|}{Philips scanner} & \multicolumn{1}{|c|}{ImagingDynamics scanner} \\
				 & \multicolumn{1}{|c||}{(N=86,460; P=306)} & \multicolumn{1}{|c|}{(N=42,823; P=110)} & \multicolumn{1}{|c||}{(N=43,625; P=196)} & \multicolumn{1}{|c|}{(N=43,047; P=270)} & \multicolumn{1}{|c|}{(N=43,413; P=36)} \\
\hline
IS ↑	&2.30	&2.21	&2.35	&\textbf{2.62}	&1.91	\\
&[2.30, 2.31]	&[2.20, 2.22]	&[2.34, 2.36]	&[2.61, 2.64]	&[1.91, 1.92]	\\
FID ↓	&\textbf{0.57}	&2.81	&2.86	&9.82	&13.17	\\
&[0.67, 0.68]	&[2.99, 3.05]	&[3.00, 3.09]	&[9.85, 10.15]	&[13.24, 13.40]	\\
Vendi Score (pixel values) ↑	&2.96	&\textbf{3.09}	&2.75	&2.87	&2.56	\\
&[2.95, 2.98]	&[3.08, 3.12]	&[2.73, 2.77]	&[2.88, 2.92]	&[2.54, 2.56]	\\
Vendi Score (hog) ↑	&\textbf{28.49}	&27.46	&21.94	&23.50	&23.29	\\
&[28.01, 28.27]	&[26.89, 27.17]	&[21.57, 21.84]	&[22.94, 23.42]	&[22.87, 22.99]	\\
Vendi Score (Inception) ↑	&5.24	&5.19	&5.07	&\textbf{6.07}	&3.96	\\
&[5.24, 5.27]	&[5.14, 5.20]	&[5.04, 5.11]	&[6.05, 6.12]	&[3.96, 3.98]	\\
\hline
RougeL (lexical) ↓	&0.08	&0.09	&0.08	&0.08	&0.09	\\
&[0.08, 0.08]	&[0.09, 0.09]	&[0.08, 0.08]	&[0.08, 0.08]	&[0.09, 0.09]  \\
Semantic diversity ↓	&0.52	&0.52	&0.52	&0.52	&0.52	\\
&[0.52, 0.52]	&[0.52, 0.52]	&[0.52, 0.52]	&[0.52, 0.52]	&[0.52, 0.52]	\\
\hline
Metadata diversity ↓	&\textbf{0.69}	&0.76	&0.83	&0.80	&0.85	\\
&[0.69, 0.69]	&[0.76, 0.76]	&[0.83, 0.83]	&[0.79, 0.80]	&[0.85, 0.85]	\\
\hline
AUC ↑	&0.796	&0.762	&\textbf{0.813}	&0.652	&0.546	\\
&$\pm$0.014	&$\pm$0.022	&$\pm$0.011	&$\pm$0.067	&$\pm$0.133	\\
\hline
\end{tabular}
}
\end{table}

%% file: table_fairness_padchest.tex
\begin{table}[htb]
\centering
\caption{Subgroups' AUC of the models trained on each subset. We report the mean AUC over the 5-fold models. $\pm$ indicates the standard deviation. \textbf{Bold} values indicate the highest mean AUC per subgroup.}
\label{tab:cxr_fairness}
\resizebox{\textwidth}{!}{%
\begin{tabular}{l|c|c||c|c|}
				  & \multicolumn{1}{|c|}{Male} & \multicolumn{1}{|c||}{Female} & \multicolumn{1}{|c|}{Philips scanner} & \multicolumn{1}{|c|}{ImagingDynamics scanner} \\
				 & \multicolumn{1}{|c|}{(N=10.798, P=50)} & \multicolumn{1}{|c||}{(N=10.812, P=27)} & \multicolumn{1}{|c|}{(N=10.684, P=74)} & \multicolumn{1}{|c|}{(N=10.931, P=3)} \\
\hline
All	model &0.794	&0.802	&0.636	&0.525	\\
&$\pm$0.025	&$\pm$0.009	&$\pm$0.031	&$\pm$0.10	\\
\hline
Male model	&\textbf{0.817}	&\textbf{0.812}	&\textbf{0.667}	&\textbf{0.693}    \\
&$\pm$0.009	&$\pm$0.018	&$\pm$0.025	&$\pm$0.093	\\
\hline
Female model	&0.759	&0.771	&0.581	&0.484	\\
&$\pm$0.025	&$\pm$0.024	&$\pm$0.037	&$\pm$0.069	\\
\hline
Philips model	&0.674	&0.616	&0.607	&0.666  \\
&$\pm$0.083	&$\pm$0.061	&$\pm$0.021	&$\pm$0.089	\\
\hline
ImagingDynamics model	&0.544	&0.551	&0.516	&0.569   \\
&$\pm$0.114	&$\pm$0.0165	&$\pm$0.037	&$\pm$0.139	\\
\hline
\end{tabular}
}
\end{table}

%% file: sec05_discussion.tex
We studied how data diversity metrics across image, text and metadata modalities correlate with each other and with classification performance on both synthetic and real datasets.  
We found limited correlations between image and metadata reference-free diversity metrics and the classification performances but higher correlations for the FID and the semantic diversity. Thanks to data maps training dynamics, we discovered a shortcut between the acquisition scanner and the disease label in the PadChest dataset. This is in line with our interview findings, where the acquisition scanner is the most important diversity factor for a clinical expert.

We acknowledge that our findings may be limited to the datasets we used, and that differences may be observed for datasets with, for example, longer radiological reports or with different class prevalence. Including more datasets in the experiments would also be beneficial to evaluate the relevance of text diversity metrics with regards to classification performances. 
Our performance analysis is focused on vision models. With the recent rise of vision-language or other multimodal models, it would be of interest to see if our findings generalise to such models. In addition, combining multiple modalities in a single metric could be of interest. While the Vendi Scores using visual features showed limited correlation, its effectiveness could potentially be improved by a similarity function using the different types of input.

Despite the limitations, our study is an important contribution to the ML community because we showcase how diversity can be evaluated from different perspectives, and the benefits such evaluation can have. In particular, it is useful to recognize shortcuts which are an increasingly recognized problem in medical imaging~\cite{jimenez2023detecting,sourget2025mask,vasquezvenegas2024detecting} but also computer vision tasks~\cite{geirhos2020shortcut}. We also show that it is not sufficient to simply assume that data is diverse, leading to better generalization - to evaluate this, we need to evaluate the metrics themselves.

To conclude, our study is grounded in combining quantitive evaluations of diversity metrics with qualitative insights from a clinical expert. We believe that beyond optimizing quantitative performance metrics, we need collaboration between domain experts and ML researchers to validate what is being measured and further explain or understand results. This is a crucial step towards the better understanding of clinical needs and challenges, leading to more effective translation of research findings into clinical practice.